\documentclass[11pt]{article}

\usepackage[final]{acl}

\usepackage{times}
\usepackage{latexsym}

\usepackage[T1]{fontenc}

\usepackage[utf8]{inputenc}

\usepackage{microtype}

\usepackage{inconsolata}

\usepackage{amsmath}
\usepackage{amssymb}
\usepackage{graphicx}
\usepackage{booktabs}
\usepackage{hyperref}
\usepackage{multicol}
\usepackage[capitalize,noabbrev]{cleveref}
\usepackage{xcolor}
\usepackage{tikz}
\usepackage{subcaption}
\usepackage[framemethod=tikz]{mdframed}
\usetikzlibrary{arrows.meta, positioning, shapes.geometric}

\definecolor{boxgray}{rgb}{.94, .94, .94}

\mdfdefinestyle{roundedbox}{%
    linecolor=gray!30,
    backgroundcolor=gray!10,
    linewidth=2pt,
    roundcorner=10pt,
    innertopmargin=8pt,
    innerbottommargin=8pt,
    innerrightmargin=8pt,
    innerleftmargin=8pt
}

\newcommand{\variabletrack}{causal variable localization track}
\newcommand{\circuittrack}{circuit localization track}

\newcommand{\mib}{\textsf{MIB}}

\newcommand{\patching}{activation patching}
\newcommand{\order}{X_{\text{Order}}}

\newcommand{\answer}{O_{\text{Answer}}}
\newcommand{\carry}{X_{\text{Carry}}}
\newcommand{\continent}{$A_{\mathit{Cont}}$}
\newcommand{\country}{$A_{\mathit{Country}}$}
\newcommand{\lang}{$A_{\mathit{Lang}}$}

\newcommand{\arxiv}[1]{#1}

\title{Findings of the BlackboxNLP 2025 Shared Task:\\Localizing Circuits and Causal Variables in Language Models}

\author{Dana Arad$^1$, Yonatan Belinkov$^{17}$, Hanjie Chen$^2$, Najoung Kim$^3$,\\
\textbf{Hosein Mohebbi$^4$, Aaron Mueller$^3$, Gabriele Sarti$^5$, Martin Tutek$^6$}  \\
  $^1$Technion -- IIT\ \ \ \ \ $^2$Rice University\ \ \ \ \ $^3$Boston University\ \ \ \ \ $^4$Tilburg University\\ $^5$University of Groningen\ \ \ \ \ $^6$University of Zagreb\ \ \ \ \ $^7$Harvard University \\}

\begin{document}
\maketitle
\begin{abstract}
Mechanistic interpretability (MI) seeks to uncover how language models (LMs) implement specific behaviors, yet measuring progress in MI remains challenging.
The recently released Mechanistic Interpretability Benchmark (\mib{}; \citealp{muellermib}) provides a standardized framework for evaluating circuit and causal variable localization.
Building on this foundation, the BlackboxNLP 2025 Shared Task extends MIB into a community-wide reproducible comparison of MI techniques.
The shared task features two tracks: circuit localization, which assesses methods that identify causally influential components and interactions driving model behavior, and causal variable localization, which evaluates approaches that map activations into interpretable features.
With three teams spanning eight different methods, participants achieved notable gains in circuit localization using ensemble and regularization strategies for circuit discovery. With one team spanning two methods, participants achieved significant gains in causal variable localization using  low-dimensional and non-linear projections to featurize activation vectors. 
The MIB leaderboard remains open; we encourage continued work in this standard evaluation framework to measure progress in MI research going forward.\footnote{\url{https://hf.co/spaces/mib-bench/leaderboard}}

\end{abstract}

\begin{figure}[t]
    \centering
    \includegraphics[width=0.95\linewidth]{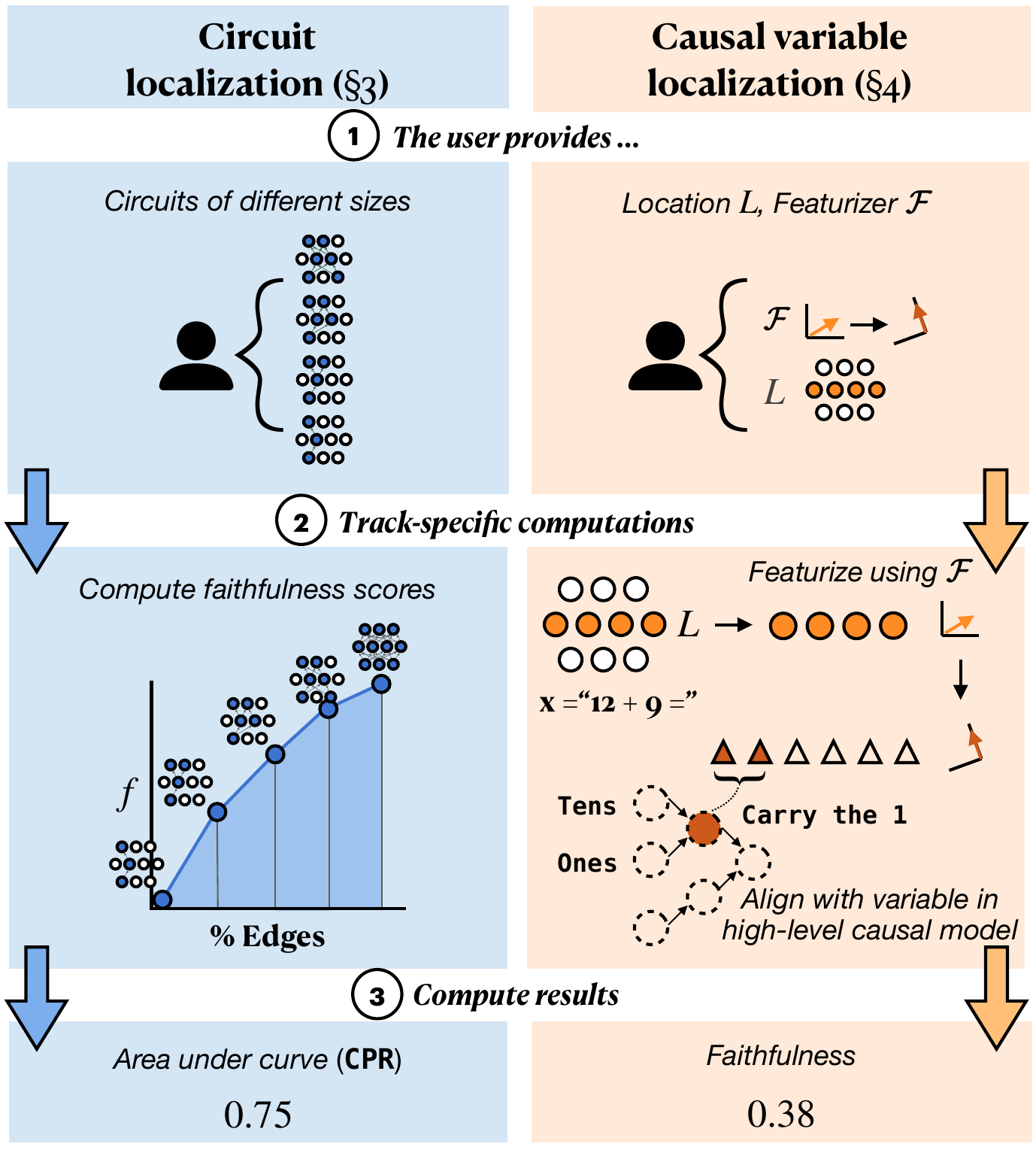}
    \caption{Overview of the evaluation method for each track in \mib{}. The circuit localization track requires uploading multiple circuits or importance scores for each component; we evaluate by taking the area under the faithfulness curve across circuit sizes. The causal variable localization track requires uploading a featurizer and location; we evaluate by intervening on the concept in the featurized space and measuring whether the model's behavior changes in the expected way. Figure reproduced from \citet{muellermib} with permission.} 
    \label{fig:overview}
    \vspace{-12pt}
\end{figure}

\section{Introduction}
The field of mechanistic interpretability (MI) is advancing rapidly, yet systematically comparing the efficacy of emerging methods remains challenging.
The recently-released Mechanistic Interpretability Benchmark (\mib{}; \citealp{muellermib}) addresses this gap by providing a standardized framework for evaluating techniques that identify circuits and localize latent causal variables in language models (LMs).
Building on this foundation, the BlackboxNLP 2025 Shared Task employs this benchmark as part of a community-wide effort aimed at accelerating progress in MI research.

The shared task comprises two tracks.
The \textbf{circuit localization track} (\S\ref{sec:circuits}) evaluates methods able to identify a minimal set of model components necessary to produce a a given behavior, such as attribution patching \citep{attribution-patching} or information flow routes \citep{ferrando-voita-2024-information}.
The \textbf{causal variable localization track} (\S\ref{sec:causalvariable}) compares methods that featurize activation vectors into more human-interpretable concepts---e.g., sparse autoencoders (SAEs; \citealp{cunningham2023sparse}) or distributed alignment search (DAS; \citealp{Geiger2024DAS}).
Submissions across these tracks are evaluated by their ability to precisely and concisely recover relevant causal pathways or causal variables in neural language models.
Submissions across both tracks are evaluated by their ability to precisely and concisely recover relevant causal pathways or causal variables in LMs.

We received submissions from four teams across the two tracks, spanning ten methods. 
Despite the relatively small number of submissions, the participating teams achieved notable performance gains across both tracks. In the circuit localization track, ensembling strategies and regularization techniques that filter components with unstable contributions to model behavior proved particularly effective, suggesting promising directions for future circuit discovery research.
In the causal variable localization Track, methods leveraging non-linear activation functions and/or multi-layer perceptrons during training demonstrated substantial improvements.

The \mib{} leaderboard will remain open for ongoing submissions to both tracks, encouraging continued participation and reproducibility.

\section{Data and Models}\label{sec:task_details}
Here, we summarize the details of \mib{}'s evaluation methods and metrics. Both tracks evaluate across four tasks representing various reasoning types, difficulty levels, and answer formats. These tasks include Indirect Object Identification (IOI), Multiple-choice Question Answering (MCQA), Arithmetic (addition and subtraction), and the AI2 Reasoning Challenge (ARC). The causal variable localization track additionally includes RAVEL \citep{huang-etal-2024-ravel}. Below, we summarize the format of each task and the size of their datasets (\S\ref{ssec:tasks}).

\subsection{Tasks}\label{ssec:tasks}
\begin{table}[t]
    \centering
    \resizebox{\linewidth}{!}{
    \begin{tabular}{lrrr}
    \toprule
    Dataset & Train & Validation & Test (Public/Private)\\
    \midrule
    IOI & 10000 & 10000 & 1000/1000\\
    MCQA & 110 & 50 & 50/50\\
    Arithmetic ($+$) & 34400 & 4920 & 1000/1000 \\
    Arithmetic ($-$) & 17400 & 2484 & 1000/1000 \\
    ARC (Easy) & 2251 & 570 & 1188/1188\\
    ARC (Challenge) & 1119 & 299 & 586/586 \\
    RAVEL & 100000 & 16000 & 1000 \\
    \bottomrule
    \end{tabular}
    }
    \caption{Dataset sizes and splits. The train, validation, and public test sets are available on \href{https://huggingface.co/mib-bench}{HuggingFace}. One may only evaluate on the private test set by uploading their circuit(s) or featurizer to the \mib{} leaderboard.}
    \label{tab:data_splits}
    \vspace{-6pt}
\end{table}

The number of instances in each dataset and split is summarized in Table~\ref{tab:data_splits}.
Each task comes with a training split on which users can discover circuits or causal variables, and a validation split on which users can tune their methods or hyperparameters. 
We also create two test sets per task: public and private. 
The public test set enables faster iteration on methods. 
We release the train, validation, and public test sets on Huggingface. 
The private test set is not visible to users; they must upload either their circuits or their featurizers to the HuggingFace leaderboard, where they are then queued for evaluation on the private test set.

\paragraph{Indirect Object Identification (IOI).}

The indirect object identification (IOI) task, first proposed by \citet{wang2022interpretability}, is one of the most studied  tasks in MI. 
IOI has sentences like \textit{``When Mary and John went to the store, John gave an apple to \_''}, containing a subject (\textit{``John''}) and an indirect object (\textit{``Mary''}), which should be completed with the indirect object. 
Even small LMs can achieve high accuracy; thus, it has been well studied \cite{cunningham2023sparse, conmy2023towards, merullo2023circuit}. 
All names tokenize to a single token for all models in \mib{}, with the private test set containing names and direct objects that are not contained in the public train or test set.

\paragraph{Arithmetic.}
Math-related tasks are common in MI \citep{stolfo2023mechanistic,nanda2023progress,zhang2024interpreting,nikankin2025arithmetic} and  interpretability research more broadly \citep{liu2023omnigrok,huang2024unified}. 
Following \citeauthor{stolfo2023mechanistic}, \mib{} defines the task as performing operations with two operands of up to two digits each.
Given a pair of numbers and an operator, the model must predict the outcome, e.g., \textit{``What is the sum of 13 and 25?"}.

\paragraph{Multiple-choice question answering (MCQA).}
MCQA is a common task format on LM evaluation benchmarks, though only a few MI works have studied it \cite{lieberum2023does, wiegreffe2024answer, li2024anchored}. 
The dataset is designed to isolate a model's MCQA ability from any task-specific knowledge \cite{wiegreffe2024answer}; the information needed to answer the questions is contained in the prompt.
Questions are about objects' colors and have four choices, such as: 
\begin{mdframed}[style=roundedbox]
\small
\begin{verbatim}
Question: A box is brown. What color is a box?
A. gray
B. black
C. white
D. brown
Answer: D
\end{verbatim}
\end{mdframed}

\paragraph{AI2 Reasoning Challenge (ARC).}
The ARC dataset \citep{clark2018think} comprises grade-school-level multiple-choice science questions. This is a representative task for evaluating basic scientific knowledge in LMs \citep{gpt3,jiang2023mistral7b,llama3}.
\mib{} follows the dataset's original partition to Easy and Challenge subsets and analyze them separately; this is due to a large accuracy difference on the two subsets.
\mib{} maintains the original 4-choice multiple-choice prompt formatting, making this dataset related in format to, but more challenging than, MCQA.

\paragraph{Resolving Attribute-Value Entanglements in Language Models (RAVEL).} RAVEL \cite{huang-etal-2024-ravel} evaluates methods for
isolating \textit{attributes} of an \textit{entity}. We include the split of RAVEL for disentangling the country, continent, and language attributes of cities. The prompts are queries about a certain attribute, e.g., \textit{Paris is on the continent of}, and the model must generate the correct completion---here, \textit{Europe}.

\subsection{Counterfactual Inputs}\label{ssec:cf}
For both \mib{} tracks, counterfactual interventions on model components form the basis for all evaluations. 
Here, components are set to the value they would take under a \textit{counterfactual input}.

In the \circuittrack, \patching\ is used to push models towards answering in an opposite manner to how they would naturally answer given the input. Success is achieved in this setting when counterfactual interventions to components outside the circuit minimally change the model's predictions.
In the \variabletrack, \patching\ is used to precisely manipulate specific concepts. Success is achieved in this setting when a variable in a causal model is a faithful summary of the role a model component plays in input-output behavior---i.e., interventions on the variable have the same effect as interventions on the model component.

\mib{} provides counterfactual inputs for each train, validation, and test samples, where the mappings from the original inputs to the counterfactual inputs are fixed to ensure consistency in evaluation.

\subsection{Models}\label{ssec:models}

\mib{} comprises of four models that cover a range of model sizes, families, capability levels, and prominence in MI:
Llama-3.1 8B \citep{llama3}, 
Gemma-2 2B \citep{team2024gemma},
Qwen-2.5 0.5B \citep{yang2024qwen2},
and GPT-2 Small \citep[117M,][]{radford2019language}.

\citet{muellermib} benchmark each model on each task and report performance. They focus specifically on model/task combinations where the model achieves at least 75\% accuracy on the task; we do the same.

\section{Circuit Localization Track}\label{sec:circuits}
The \circuittrack{} centers on evaluating how well a method can discover causal subgraphs $\mathcal{C}$ of a computation graph; these are more commonly known as \textbf{circuits} \citep{olah2020zoom}. The purpose of circuits is to localize the mechanisms underlying how a full neural network $\mathcal{N}$ performs a given task. A circuit $\mathcal{C}$ is a graph consisting of nodes and edges between components in $\mathcal{N}$. Nodes are typically submodules or attention heads (e.g., the layer 5 MLP, or attention head 10 at layer 12); edges reflect information flow between a pair of nodes.

A typical circuit discovery pipeline consists of two stages: (1) scoring the full set of graph components (nodes, edges, etc.), and (2) selecting a subset of the components that constitute the circuit.

\subsection{Metrics}\label{sec:metrics} 
\mib{} defines two circuit localization metrics: the \arxiv{\textbf{integrated circuit performance ratio} (\textsf{CPR}), and the \textbf{integrated circuit-model distance} (\textsf{CMD})}.
\textsf{CPR} measures whether a series of circuits include components with a positive effect on model performance on the task; higher is better.
\textsf{CMD} measures whether a series of circuits yield \emph{the same} strength of preference for the correct answer as the full model; 0 is best, and corresponds to no difference between the circuit and full model behavior with respect to predicting the correct answer. 
Intuitively, \textsf{CPR} may be more useful for finding circuits that cause the model to perform well on the task, while \textsf{CMD} may be more useful when the aim is to explain the full algorithm the model implements to perform some behavior (including cases where the behavior is not desirable).

Given a circuit $\mathcal{C}$ and the full model $\mathcal{N}$, faithfulness $f$ is defined as:
\begin{equation}
    f(\mathcal{C}, \mathcal{N}; m) = \frac{m(\mathcal{C}) - m(\varnothing)}{m(\mathcal{N}) - m(\varnothing)},
    \label{eq:f}
\end{equation}
where $m$ is the logit difference $y' - y$ between the correct answer $y$ given the original input $x$ and correct answer $y'$ given the counterfactual input $x'$.

Thus, \textsf{CPR} is computed as the area under the faithfulness curve with respect to circuit size. Following \citet{muellermib}, we approximate this area using a Riemann sum over $f$ computed across circuit sizes. \textsf{CMD} as the area between the faithfulness curve and 1; we also approximate this using a Riemann sum.

\paragraph{Measuring circuit size.}
\mib{} treats including a node as equivalent to including all of its outgoing edges, and including one neuron\footnote{We use ``neuron'' to refer to a single dimension of any hidden vector, regardless of whether it is preceded by a non-linearity.} 
of $d_\text{model}$ in submodule $u$ as including all outgoing edges from $u$ to $\frac{1}{d_\text{model}}$ of the degree they would have been compared to including all neurons in $u$.

Under these assumptions, \mib{} defines the weighted edge count:
\begin{equation}
|\mathcal{C}| = \sum_{(u,v) \in \mathcal{C}} \left(\frac{|N_{u} \cap N_{\mathcal{C}}|}{|N_u|}\right),
\end{equation}
where $u$ and $v$ are nodes (submodules), $N_u$ is the set of neurons in $u$ (the size of which is typically $d_\text{model}$), and $N_\mathcal{C}$ is the set of neurons in the circuit.
This count is then normalized by the number of possible edges to obtain a percentage.

\subsection{Submission Procedure}
All results below are computed on the private test splits for each task. To evaluate on the private test split, participants were first required to upload their circuits to a HuggingFace repository.\footnote{See \href{https://huggingface.co/mib-bench/mib-circuits-example/tree/main/importances/pt}{this repository} for an example of how circuit repositories were required to be structured.} The faithfulness evaluation required 9 circuits of different sizes; we expected one circuit $\mathcal{C}_k$ for each $k\in K$, where $k$ is the maximum proportion of components in $\mathcal{N}$ that are allowed to remain in the circuit. Here, $K=\{0.001,0.002,0.005,0.01,0.02,0.05,0.1,0.2,0.5\}$. For each model/task combination, a folder of circuits was required. Each circuit is a dictionary, where each node and edge is a key whose value is either a boolean indicating whether the component or edge belongs to the circuit, or a floating-point importance score. If the user uploaded floating-point importance scores, then only one file per model/task was required; we took the top-$k$ components by importance for each circuit size $k\in K$. If the user uploaded binary inclusion indices, they were required to upload one circuit file for each threshold $k\in K$.

Users provided a link to this repository on the ``Submit'' tab of the \mib{} leaderboard,\footnote{\url{https://hf.co/spaces/mib-bench/leaderboard}} along with a method name.

\subsection{Task Submissions}
We received submissions from three teams for the circuit localization track covering eight proposed methods. We taxonomize and summarize the approaches here.

\begin{table*}[t]
    \centering
    \resizebox{\linewidth}{!}{
    \begin{tabular}{@{}l@{} @{\hspace{7pt}}r @{\hspace{7pt}}r @{\hspace{5pt}}r @{\hspace{5pt}}r @{\hspace{5pt}}r r r @{\hspace{5pt}}r @{\hspace{5pt}}r r @{\hspace{5pt}}r r}
    \toprule
    & \multicolumn{5}{c}{IOI} & Arithmetic & \multicolumn{3}{c}{MCQA} & \multicolumn{2}{c}{ARC (E)} & ARC (C) \\\cmidrule(lr){2-6}\cmidrule(lr){7-7}\cmidrule(lr){8-10}\cmidrule(lr){11-12}\cmidrule(lr){13-13}
    \textbf{Method} & InterpBench ($\uparrow$) & GPT-2 & Qwen-2.5 & Gemma-2 & Llama-3.1 & Llama-3.1  & Qwen-2.5 & Gemma-2 & Llama-3.1 & Gemma-2 & Llama-3.1 & Llama-3.1 \\
    \midrule
    Random & 0.44 & 0.75 & 0.72 & 0.69 & 0.74 & 0.75 & 0.73 & 0.68 & 0.74 & 0.68 & 0.74 & 0.74 \\
    \midrule
    EAP (mean) & \underline{0.78} & 0.29 & 0.18 & 0.25 & \underline{0.04} & 0.07 & 0.21 & 0.20 & 0.16 & \underline{0.22} & \underline{0.28} & \underline{0.20} \\
    EAP (CF) & 0.73 & \underline{0.03} & 0.15 & 0.06 & \textbf{0.01} &  \underline{0.01} & 0.07 & 0.08 & \textbf{0.09} & \underline{0.04} & \textbf{0.11} & \textbf{0.18} \\
    EAP (OA) &  0.77 & 0.30 & 0.16 & - & - & - & 0.11 & - & - & - & - & -  \\
    \midrule
    EAP-IG-inp. (CF) & 0.71 & \underline{0.03} & \underline{0.02} & \underline{0.04} & \textbf{0.01} & \textbf{0.00} & 0.08 & \textbf{0.06} & 0.14 & \underline{0.04} & \textbf{0.11} & 0.22 \\
    EAP-IG-act. (CF) & \textbf{0.81}  & \underline{0.03} & \textbf{0.01} & \textbf{0.03} & \textbf{0.01} &  \textbf{0.00} & \underline{0.05} & \underline{0.07} & \underline{0.13} & \underline{0.04} & 0.30 & 0.37 \\
    \midrule
    P-Ens \cite{mondorf2025blackboxnlp} & - & \textbf{0.02} & \underline{0.02} & - & - & - & 0.07 & - & - & - & - & - \\
    S-Ens \cite{mondorf2025blackboxnlp} & - & \underline{0.03} & \underline{0.02} & - & - & - & 0.07 & - & - & - & - & - \\
    Hybrid-Ens \cite{mondorf2025blackboxnlp} & - & \underline{0.03} & \underline{0.02} & - & - & - & \textbf{0.04} & - & - & - & - & - \\
    ILP + PNR + Bootstrapping (\citeyear{nikankin2025blackboxnlp}) & - & \textbf{0.02} & \textbf{0.01} & \underline{0.04} & \textbf{0.01} & \underline{0.01} & 0.08 & \underline{0.07}  & 0.45 & \textbf{0.03} & - & - \\
    IPE (CF) \cite{brunello2025blackboxnlp} & - & \textbf{0.02}  & 0.57 & - & - & 0.54 & - & - & - & - & - & - \\
    \bottomrule
    \end{tabular}}
    \caption{\textsf{CMD} scores across circuit localization methods (lower is better) on the private test set. All evaluations were performed using counterfactual ablations. Arithmetic scores are averaged across addition and subtraction. We \textbf{bold} and \underline{underline} the best and second-best methods per column, respectively.}
    \label{tab:subgraph_cmd}
    \vspace{-3pt}
\end{table*}

\begin{table*}[t]
    \centering
    \resizebox{\linewidth}{!}{
    \begin{tabular}{lrrrrrrrrrrr}
    \toprule
    & \multicolumn{4}{c}{IOI} & Arithmetic & \multicolumn{3}{c}{MCQA} & \multicolumn{2}{c}{ARC (E)} & ARC (C) \\\cmidrule(lr){2-5}\cmidrule(lr){6-6}\cmidrule(lr){7-9}\cmidrule(lr){10-11}\cmidrule(lr){12-12}
    \textbf{Method} & GPT-2 & Qwen-2.5 & Gemma-2 & Llama-3.1 & Llama-3.1  & Qwen-2.5 & Gemma-2 & Llama-3.1 & Gemma-2 & Llama-3.1 & Llama-3.1 \\
    \midrule
    EActP (CF) & \textbf{2.30} & 1.21 & - & - & - & 0.85 & - & - & - & - & - \\
    \midrule
    EAP (mean) &  0.29 & 0.71 & 0.68 & 0.98 & 0.35 & 0.29 & 0.33 & 0.13 & 0.26 & 0.34 & 0.80 \\
    EAP (CF) & 1.20 & 0.26 & 1.29 & 0.85 & 0.55 & 0.85 & 1.49 & 1.00 &
    1.08 & \underline{0.80} & \underline{0.82} \\
    EAP (OA) & 0.95 & 0.70 & - & - & - & 0.29 & - & - & - & - & - \\
    \midrule
    EAP-IG-inputs (CF) & 1.85 & 1.63 & \textbf{3.20} & 2.08 & \underline{0.99} & \underline{1.16} & 1.64 & 1.05 & 1.53 & \textbf{1.04} & \textbf{0.98} \\
    EAP-IG-activations (CF) &  1.82 & 1.63 & 2.07 & 1.60 & 0.98 & 0.77 & 1.57 & 0.79 & \textbf{1.70} & 0.71 & 0.63 \\
    \midrule
    NAP (CF) &  0.28 & 0.30 & 0.30 & 0.26 & 0.27 & 0.38 & 1.47 & \underline{1.69} & 1.01 & 0.26 & 0.26 \\
    NAP-IG (CF) &  0.76 & 0.29 & 1.52 & 0.42 & 0.39 & 0.77 & \textbf{1.71} & \textbf{1.87} & 1.53 & 0.26 & 0.26 \\
    \midrule
    P-Ens \cite{mondorf2025blackboxnlp} & 2.11 & \textbf{1.88} & - & - & - & 0.79 & - & - & - & - & - \\
    S-Ens \cite{mondorf2025blackboxnlp} & \underline{2.37} & \underline{1.71} & - & - & - & \underline{1.16} & - & - & - & - & - \\
    Hybrid-Ens \cite{mondorf2025blackboxnlp} & \textbf{2.43} & \textbf{1.88 }& - & - & - & \textbf{1.19} & - & - & - & - & - \\
    ILP + PNR + Bootstrapping (\citeyear{nikankin2025blackboxnlp}) & 1.89 & \underline{1.71} & \underline{3.01} & \textbf{2.39} & \textbf{1.04} & 1.04 & \underline{1.7}  & 1.22 & \underline{1.63} & - & - \\
    IPE (CF) \cite{brunello2025blackboxnlp} & 2.24  & 0.35 & - & - & - & 0.45 & - & - & - & - & - \\
    \bottomrule
    \end{tabular}}
    \caption{\textsf{CPR} scores across circuit localization methods on the private test set. All evaluations were performed using counterfactual ablations. Higher scores are better. Arithmetic scores are averaged across addition and subtraction. We \textbf{bold} and \underline{underline} the best and second-best methods per column, respectively.}
    \label{tab:subgraph_cpr}
\end{table*}

\paragraph{Ensemble scoring strategies.} \citet{mondorf2025blackboxnlp} proposed ensembling two or more circuit localization methods to improve attribution scores. They examined three ensembling variants: \textit{parallel}, \textit{sequential},  and their \textit{hybrid} combination.

\textit{Parallel ensembling (P-Ens)} merges the scores from different methods into a single edge, using scores from the three variants of edge patching implemented by \citeauthor{muellermib}: (1) Edge Attribution Patching (EAP; \citealp{attribution-patching, syed-etal-2024-attribution}, (2) EAP-IG-inputs \cite{hanna2024have}, and (3) EAP-IG-activations \cite{marks2024sparsefeaturecircuitsdiscovering}. The latter two methods complement EAP with integrated gradients~\citep{sundararajan2017ig} to improve estimates of edge importance, perturbing input embeddings and activations, respectively.
The authors experimented with score merging using mean, weighted average, maximum, and minimum, and found that mean yielded the best results.

\textit{Sequential ensembling (S-Ens)} utilizes attribution scores produced by a fast circuit identification method to warm-start a slower, more precise method, thereby achieving faster convergence and further refining the initial scores.
Specifically, they use EAP-IG-inputs \cite{hanna2024have} edge attribution values to initialize the learnable log alpha parameters of edge pruning \cite{bhaskar2024finding}.

Finally, \textit{hybrid ensembling (Hybrid-Ens)} combines the parallel and sequential strategies by taking the unweighted average over all four method--the three EAP variants and warm-start edge pruning for all model-task combination.

\paragraph{Improved edge selection.} Focusing on the second stage of the circuit discovery pipeline, \citet{nikankin2025blackboxnlp} experimented with three methods to improve edge selection process.
First, they observe that EAP-IG scores can vary across data samples from the same task, with some edges receiving both negative and positive values in different samples. 
The score sign is significant, as it signifies whether the edge contributes positively or negatively to the performance on the task.
By bootstrapping the scores across resamples of the training data, they identify edges with consistent score signs and filter out unstable ones.

Second, they introduce a ratio based strategy for edge selection based on their signs (PNR): select a fixed proportion of top positive edges, and the rest by absolute value. This approach allows finer control over the balance of
edge types and improves circuit faithfulness.
Lastly, they formulate circuit construction as an Integer Linear Programming (ILP) optimization problem, instead of using the naive greedy solution.

\paragraph{Path scoring.} \citet{brunello2025blackboxnlp} proposed Isolating Path Effects (IPE) to identify entire computational paths from input embeddings to output logits responsible for certain model behaviors, as opposed to individual edges.
Their method modifies the messages passed between nodes along a given path in such a way as to either precisely remove the effects of the entire path (i.e., ablate it) or to replace the path’s effects with those that a counterfactual input would have produced.
IPE differs from current path-patching or edge-activation-patching techniques, as they do not ablate individual paths but rather a set of paths sharing certain edges, thereby allowing a more precise tracing of information flow.

\subsection{Results}
Table~\ref{tab:subgraph_cmd} and Table~\ref{tab:subgraph_cpr} show the \textsf{CMD} and \textsf{CPR} scores, respectively, of the top method from each submission as well as selected methods from \mib{}, on the private test set. All submissions perform especially well, achieving better or comparable scores to even the strongest baselines. 

The submission of \citet{nikankin2025blackboxnlp} achieves especially strong \textsf{CMD} scores, whereas the Hybrid-Ens method of \citet{mondorf2025blackboxnlp} achieves the strongest CPR scores. The IPE method by 
\citet{brunello2025blackboxnlp} also performs well on IOI for GPT-2.
Among the methods of \citet{mondorf2025blackboxnlp}, Hybrid-Ens performs the strongest across tasks. These results suggest that ensembling strategies may be an accessible and fruitful line of work for future circuit discovery research.
For \citet{nikankin2025arithmetic}, the removal of components with inconsistent effects on model outputs and a mixture of positive and high-magnitude components may have a regularizing effect on the discovered circuit, causing it to behave more closely to the whole model and potentially suppressing components that would have strong but inconsistent impacts on model behavior. It would be interesting to see detailed comparisons of each method on more fine-grained distributions to characterize when and why each is likely to succeed. That said, there is no clear winner; the best method appears to depend on the chosen metric.

A factor we have not directly evaluated for is the time complexity of each method. It is possible that different methods could perform comparably despite having very different expected runtimes; a direct comparison of compute requirements would be valuable in helping future researchers decide which methods are most worthwhile to run. We note that many cells are missing for each submission, but this does not necessarily reflect compute requirements---this could be due to local memory constraints, runtime limitations, or other compute constraints (e.g., limited access to GPUs on a cluster before a deadline).

\section{Causal Variable Localization Track}\label{sec:causalvariable}
The causal variable localization track focuses on evaluating how well a method can discover specific causal variables in a language model's activation space. The basic intuition is that any hidden vector $\mathbf{h} \in \mathbb{R}^d$ constructed by a model $\mathcal{N}$ during inference can be mapped into a new feature space $\mathbb{F}^k$ (e.g., a rotated vector space) using an invertible function $\mathcal{F}: \mathbb{R}^d \to \mathbb{F}^k$ (e.g., multiplication with an orthogonal matrix).
Features $\Pi$ are a set of indices between 1 and $k$, i.e., a set of dimensions in $\mathbb{F}^k$. 
This framework supports features like neurons, orthogonal directions, (sets of) SAE features, and non-linear features. The vector $\mathbf{h}$ might come from the residual stream between transformer layers or the output of an attention head.

\subsection{Evaluation Metric}
We use \textbf{faithfulness} to evaluate causal variable localization submissions. This metric captures the degree to which the provided features capture the causal variable under counterfactual intervention. To evaluate faithfulness, we use \textit{interchange interventions}. Given base and counterfactual inputs $(b,c)$, high-level causal graph $\mathcal{H}$, and causal variable $X\in\mathcal{H}$, the interchange intervention $\mathcal{H}_{X \leftarrow \mathsf{Get}(\mathcal{H}(c), X)}(b)$ runs $\mathcal{H}$ on base input $b$ while fixing the variable $X$ to the value it takes when $\mathcal{H}$ is run on a counterfactual input $c$ \cite{vig2020, geiger-etal-2020-neural}. The distributed interchange intervention $\mathcal{N}_{\Pi_X \leftarrow \mathsf{Get}(\mathcal{N}({c}), \Pi_X)}(b)$ runs $\mathcal{N}$ on $b$ while fixing the features $\Pi_X$ of the hidden vector $\mathbf{h}$ passed through $\mathcal{F}$ to the value they take for counterfactual input $c$ \cite{ Wu2023, amini-etal-2023-naturalistic, Geiger2024DAS}. Faithfulness is measured as the proportion of examples for which the intervention yields the expected change in the model's output behavior. See \citet{wu2023interpretability} and \citet{muellermib} for examples.

\begin{table*}[t]
\small
\centering
\begin{subfigure}[b]{0.6\textwidth}
\centering
  \resizebox{\textwidth}{!}{
  \begin{tabular}{lrrrrrr}
    \toprule
     & \multicolumn{6}{c}{RAVEL} \\
    \cmidrule(lr){2-7}
 & \multicolumn{3}{c}{Gemma-2} & \multicolumn{3}{c}{Llama-3.1} \\
    \cmidrule(lr){2-4}\cmidrule(lr){5-7}
    Method & \continent & \country & \lang& \continent & \country & \lang\\
    \midrule
    DAS & 75 (\textbf{85}) & 57 (\textbf{67}) & 62 (\textbf{70}) & 75 (\textbf{83}) & 58 (\textbf{64}) & 63 (\textbf{70}) \\
    DBM & 66 (\textbf{71}) & 53 (\textbf{65}) & 54 (\textbf{58}) & 68 (\textbf{80}) & 53 (\textbf{59}) & 58 (\textbf{64}) \\
    +PCA & 63 (\textbf{70}) & 47 (\textbf{53}) & 50 (\textbf{56}) & 62 (\textbf{74}) & 48 (\textbf{54}) & 53 (\textbf{57}) \\
    +SAE & 64 (\textbf{72}) & 49 (\textbf{56}) & 53 (\textbf{59}) & 64 (\textbf{72}) & 50 (\textbf{57}) & 55 (\textbf{57}) \\
    Full Vector & 48 (\textbf{62}) & 49 (\textbf{57}) & 45 (\textbf{56}) & 53 (\textbf{62}) & 47 (\textbf{53}) & 47 (\textbf{57}) \\
    \midrule
    Orthogonal & - & - & - & 84 (\textbf{89}) & 70 (\textbf{79}) & 72 (\textbf{79}) \\
    Nonlinear & - & - & - & 83 (\textbf{89}) & 70 (\textbf{78}) & 72 (\textbf{79}) \\
    \bottomrule
  \end{tabular}}
  \caption{The RAVEL task with variables for the country \country, continent \\ \continent, and language \lang\ of a city.}\vspace{10pt}
  \label{tab:results-ravel}
\end{subfigure}
\hfill
\begin{subfigure}[b]{0.37\textwidth}
    \centering
  \begin{tabular}{lrr}
    \toprule
     & \multicolumn{2}{c}{Arithmetic (+)} \\
    \cmidrule(lr){2-3}
 & \multicolumn{1}{c}{Gemma-2} & \multicolumn{1}{c}{Llama-3.1} \\
    \cmidrule(lr){2-2}\cmidrule(lr){3-3}
    Method & $\carry$ & $\carry$ \\
   \midrule
    DAS & 31 (\textbf{35}) & 54 (\textbf{65}) \\
    DBM & 33 (\textbf{43}) & 47 (\textbf{58}) \\
    +PCA & 32 (\textbf{44}) & 37 (\textbf{56}) \\
    +SAE & 32 (\textbf{44}) & 38 (\textbf{55}) \\
    Full Vector & 29 (\textbf{35}) & 35 (\textbf{45}) \\
    \midrule
    Orthogonal & - & 53 (\textbf{65}) \\
    Nonlinear & - & - \\
    \bottomrule
  \end{tabular}
  \caption{The two-digit arithmetic task with a variable computing the carry-the-one operation ($\carry$).}\vspace{10pt}
  \label{tab:results-arithmetic}
\end{subfigure}

\begin{subfigure}[b]{0.6\textwidth}
\centering
  \centering
  \begin{tabular}{lrrrrrr}
    \toprule
     & \multicolumn{6}{c}{MCQA} \\
    \cmidrule(lr){2-7}
 & \multicolumn{2}{c}{Gemma-2} & \multicolumn{2}{c}{Llama-3.1} & \multicolumn{2}{c}{Qwen-2.5} \\
    \cmidrule(lr){2-3}\cmidrule(lr){4-5}\cmidrule(lr){6-7}
    Method & $\answer$ & $\order$ & $\answer$ & $\order$ & $\answer$ & $\order$ \\
    \midrule
    DAS & 95 (\textbf{97}) & 77 (\textbf{93}) & 94 (\textbf{100}) & 77 (\textbf{91}) & 86 (\textbf{95}) & 78 (\textbf{100}) \\
    DBM & 84 (\textbf{99}) & 63 (\textbf{84}) & 86 (\textbf{100}) & 66 (\textbf{73}) & 46 (\textbf{94}) & 60 (\textbf{99}) \\
    +PCA & 57 (\textbf{96}) & 52 (\textbf{81}) & 65 (\textbf{99}) & 53 (\textbf{74}) & 22 (\textbf{76}) & 54 (\textbf{100}) \\
    +SAE & 73 (\textbf{90}) & 51 (\textbf{65}) & 80 (\textbf{99}) & 58 (\textbf{65}) & -- & -- \\
    Full Vector & 61 (\textbf{100}) & 44 (\textbf{77}) & 77 (\textbf{100}) & 46 (\textbf{68}) & 35 (\textbf{99}) & 49 (\textbf{99}) \\
    \midrule
    Orthogonal & - & - & - & - & 90 (\textbf{98}) & 78 (\textbf{100}) \\
    Nonlinear & - & - & 95 (\textbf{100}) & 81 (\textbf{94}) & 89 (\textbf{98}) & 81 (\textbf{100}) \\
    \bottomrule
  \end{tabular}
  \caption{The MCQA task with variables for the ordering of the answer $\order$ and then the answer token $\answer$. This is a low-data regime ($\approx$100 examples).}
  \label{tab:results-mcqa}
\end{subfigure}
\caption{Results for the \variabletrack. Table headers show the task, the model, and the selected causal variable, respectively. 
We do not report results for ARC or IOI, as no submissions were made for these tasks.  We report interchange intervention accuracy (i.e., our faithfulness metric), i.e., the proportion of aligned interventions on the causal model and deep learning model that result in the same output token(s); higher is better. For each method of aligning a causal variable to LM features, we report the mean across counterfactual datasets and layers in the low-level model. In parenthesis and \textbf{bold}, we report the best alignment across all layers.}
\label{tab:causalgraph}
\end{table*}

\subsection{Submission Procedure}
As for the circuit localization task, users were required to upload files to a HuggingFace repository, although the required files differed for causal variable localization.\footnote{See \href{https://huggingface.co/mib-bench/mib-causalvariable-example/tree/main}{this repository} for an example of how causal variable localization repositories were required to be structured.} Here, a user was required to upload at least three artifacts for a given causal variable: a trained featurizer $\mathcal{F}$, a trained inverse featurizer $\mathcal{F}^{-1}$, and position indices corresponding to the dimensions of the featurized space that encode the causal variable of interest. If the featurizer was not one of the supported baseline types, users were also required to upload Python code that could save and load their featurizer. We also supported interventions at dynamic token positions; if used, users were required to upload a Python script specifying which token positions to intervene on for a given example.\footnote{See the track's \href{https://github.com/atticusg/CausalAbstraction/tree/main}{GitHub repository} for further details.}

\subsection{Task Submissions}
We received submissions from one team totalling two methods \citep{hirlimann2025blackboxnlp}. Both methods extend the official Distributed Alignment Search (DAS; \citealp{Geiger2024DAS}) baseline.

\paragraph{Non-linear featurizer.} This method extends DAS with a multi-layer perceptron (MLP) and non-linearities. During training, this method augments the feature mixing stage with an MLP:
\begin{align}
    \mathbf{h} &= \text{GeLU}(W_u\mathbf{x})\\
    \mathbf{\hat{x}} &= \text{tanh}(W_d\mathbf{h}),
\end{align}
where $W_u$ and $W_d$ are learned up-projection and down-projection weights, respectively. This is only applied during training. This allows the featurizer to ``blend'' potentially independent representations and go beyond convex combinations of features, which could allow it to learn dependencies where the signal is not strictly separable by individual directions in the original actuvation space $\mathbf{x}$. Empirically, this was the most well-performing method, even outperforming DAS---the best-performing baseline method. That said, recent work has demonstrated that non-linear featurizers are highly expressive, and as such can locate potentially any feature, including those that are not in the model itself \citep{sutter2025nonlinearrepresentationdilemmacausal}, echoing the memorization problem that characterized probing classifiers \citep{belinkov-2022-probing}. Additional validation is needed to confirm that the learned features capture genuine variables the model employs during processing.

\paragraph{Orthogonal non-linear projection.} This featurizer is a simplified variant of the non-linear featurizer. Here, the features pass only through a tanh non-linearity without a feed-forward layer. This still enables rich feature interactions to be learned, but does not have as much expressive power as the non-linear featurizer.


\subsection{Results}
We show faithfulness scores for baselines and submissions in Table~\ref{tab:causalgraph}. Both the orthogonal and non-linear methods achieve significant gains over DAS across tasks and models. Despite the greater expressive power of the non-linear featurizers, this method performs comparably to the simpler orthogonal featurizer across tasks, with non-linear featurization proving slightly stronger for MCQA with Qwen-2.5.

\section{Conclusions}
Despite the relatively small number of submissions, participants achieved significant performance gains. Ensembling methods are quite effective for circuit discovery, as is regularization via filtering components with unstable contributions to model behavior; we encourage future work to continue exploring these directions. Furthermore, one can achieve significant gains in variable localization using non-linear mediator types; these projections into new spaces can be highly effective with the proper training procedure, even when the non-linearity is built on top of a simple architecture. This suggests that expressive featurizer training formulations that leverage existing mediator types might yield significant gains in causal variable localization---but more controls are needed to ensure that concepts truly in the model itself are being isolated (as opposed to the featurizer learning the causal variable itself).

The \mib{} leaderboard will continue to accept public submissions in both tracks. The results of this shared task will inform the experimental design and baseline choices for future studies employing circuits and causal-variable localization methods in language models. We hope that participants will continue to publicize their findings to benefit the community and enable scientific progress through direct comparisons in a shared-task setting.

\section*{Acknowledgments}
D.A. is supported by the
Ariane de Rothschild Women Doctoral Program. 
G.S. acknowledges the support of the Dutch Research Council (NWO) for the project InDeep (NWA.1292.19.399).
This research was partly supported by an Azrieli Foundation
Early Career Faculty Fellowship,  Open Philanthropy, a Google Award, and the European Union (ERC, Control-LM, 101165402). Views and opinions expressed are however
those of the author(s) only and do not necessarily reflect those of the European Union or the European
Research Council Executive Agency. Neither the European Union nor the granting authority can be
held responsible for them.
\bibliography{custom}




\end{document}